\documentclass[letterpaper,10pt,journal,twoside]{IEEEtran}

\IEEEoverridecommandlockouts                              % This command is only needed if 
\usepackage{times}
\usepackage{graphicx}
\usepackage{subfigure}
\usepackage{capt-of}
\usepackage{amsmath,amssymb,amsopn,amstext,amsfonts,amsthm}
\usepackage{cancel}
\usepackage[space]{cite}
\usepackage{pdfsync}
\usepackage{balance}
\usepackage{color}
\usepackage{mathtools}
\usepackage{bm}
\usepackage{diagbox}
\usepackage{float}
\usepackage{epstopdf}
\usepackage{pifont}
\usepackage{fixltx2e}
\usepackage{multirow}
\usepackage{url}
\usepackage{verbatim}
\usepackage[linkcolor=black,citecolor=black,urlcolor=black,colorlinks=true]{hyperref}
\usepackage{soul,xcolor}
\usepackage[linesnumbered,ruled,vlined]{algorithm2e}
\usepackage{lipsum}
\usepackage{booktabs}
\usepackage{subfiles}
\usepackage{array,booktabs}
\usepackage{mwe}
\usepackage{tablefootnote}
\newcolumntype{M}[1]{>{\centering\arraybackslash}m{#1}}

\graphicspath{{./img/}} 
\DeclareGraphicsExtensions{.png,.jpg,.eps,.pdf,.jpeg}

\title{UniQuad: A Unified and Versatile Quadrotor Platform Series for UAV Research and Application}
\author{Yichen Zhang$^\dagger$, Xinyi Chen, Peize Liu, Junzhe Wang, Hetai Zou, Neng Pan, Fei Gao and Shaojie Shen%
\thanks{$^\dagger$ Corresponding author: Yichen Zhang

This work was supported by HKUST Postgraduate Studentship and HKUST-DJI Joint Innovation Laboratory.

Y. Zhang, X. Chen, P. Liu, J. Wang, H. Zou are with the Department of Electronic and Computer Engineering, Hong Kong University of Science and Technology, Hong Kong, China.

N. Pan, F. Gao are with the Institute of Cyber-Systems and Control, College of Control Science and Engineering, Zhejiang University, Hangzhou, China, and also with Huzhou Institute, Zhejiang
University, Huzhou, China.
\tt\footnotesize $\{$yzhangec,xchencq,pliuan,jwanggj,hzouah,eeshaojie$\}$\\@ust.hk, $\{$panneng\_zju, fgaoaa$\}$@zju.edu.cn
}%
}

\begin{document}
\maketitle

% \vspace{-1.6cm}

\begin{abstract}
% 65 - 100 words
As quadrotors take on an increasingly diverse range of roles, researchers often need to develop new hardware platforms tailored for specific tasks, introducing significant engineering overhead. In this article, we introduce the \textbf{UniQuad} series, a unified and versatile quadrotor platform series that offers high flexibility to adapt to a wide range of common tasks, excellent customizability for advanced demands, and easy maintenance in case of crashes. This project is fully open-source at \url{https://hkust-aerial-robotics.github.io/UniQuad}.
\end{abstract}

\begin{IEEEkeywords}
Aerial Systems: Perception and Autonomy; Aerial Systems: Applications; Engineering for Robotic Systems
\end{IEEEkeywords}

\section{Introduction}
\label{sec:intro}
% 140 - 180 words

% [What is the question? Why is it important and worth studying? Background and motivation]
% [How others have approached this problem? What are the limitations of their approach?]
Quadrotors are taking charge of an increasingly diverse range of tasks in both research topics and industrial applications, such as autonomous exploration \cite{zhang2024falcon}, 3D reconstruction \cite{feng2023fc} and aerial delivery \cite{li2023autotrans}.
% autonomous exploration \cite{cao2021tare,zhou2021fuel}, 3D reconstruction \cite{song2021view, feng2023fc} and aerial delivery \cite{foehn2017fast, li2023autotrans}.
Different tasks may have varying requirements for the quadrotor platform in terms of equipped sensors, battery life, quadrotor size and payload capability.

Existing open-source quadrotor projects are mostly targeted at a specific emphasis and suitable for only a limited range of tasks.
For example, early platforms such as ASL-Flight \cite{sa2017build}, FLA \cite{mohta2018fast} and MRS \cite{baca2021mrs} for autonomous flight tend to have a rather large size with a wheelbase around $500$mm.
This prohibits their usage in cluttered environments and limits the agent's agility.
Agilicious \cite{foehn2022agilicious} provides a high thrust-to-weight platform for agile flight tasks.
The compactness achieved through a cramped hardware design restricts it to vision-based tasks using only visual sensors.
Platforms with smaller dimensions \cite{zhou2022swarm,pan2023canfly,liu2024omninxt} may not support heavier sensors like LiDAR or ferry additional payloads.
As a result, research groups frequently need to develop new hardware platforms tailored for the specific task, which introduces significant engineering overhead and impedes research progress.

\begin{figure}[t]
	\centering
  % \vspace{0.4cm}
  \includegraphics[width=\columnwidth]{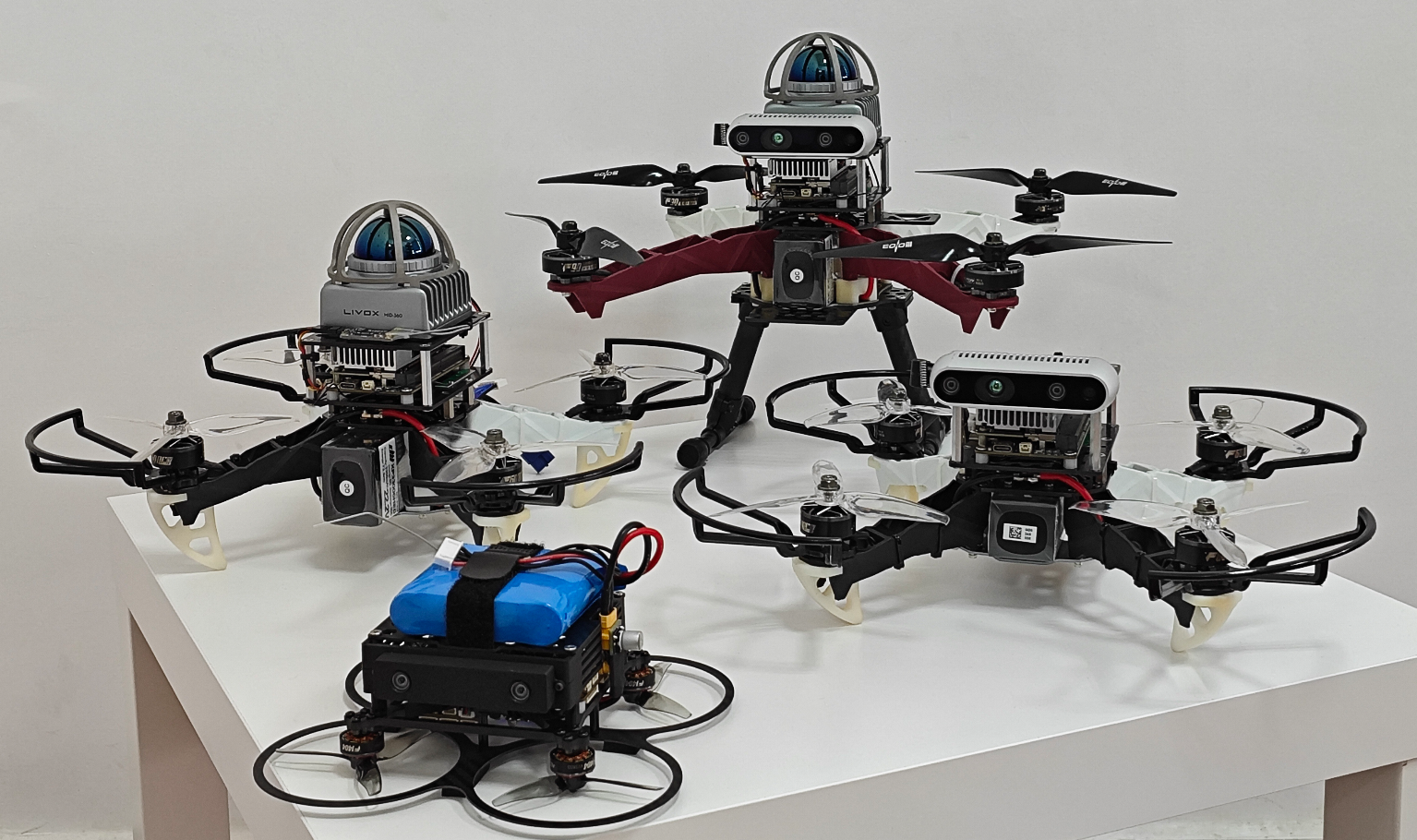}  
%   \vspace{-0.6cm}
  \caption{\label{fig:all_uav_pic} The quadrotor platforms in \textbf{UniQuad} series provided in this article, namely \textbf{Uni250L}, \textbf{Uni127C}, \textbf{Uni350CL} and \textbf{Uni250C} from left to right.}
  \vspace{-1.6cm}
\end{figure}

% [How we solved the problem? Motivations?]
With the awareness of the growing diversity of quadrotor applications, we introduce a unified and versatile quadrotor platform series \textbf{UniQuad} that shares a common conceptual design with modular components.
% This offers high flexibility to adapt to a wide range of common tasks, good customizability for advanced demands, and easy maintenance in case of crashes.
Note that the four models depicted in Fig.\ref{fig:all_uav_pic} merely serve as examples of the platform series, which can be further extended and customized on your own choice.
This platform series aims to alleviate engineering overhead for researchers working on diverse UAV topics and encourage reproducible real-world verification and application.

% [Describe the results and summarize contribution bullets]

% \subfile{sections/related.tex}

% \begin{figure}[t]
% 	\centering
%   \includegraphics[width=0.5\columnwidth]{cat_think.jpg}  
% %   \vspace{-0.6cm}
%   \caption{\label{fig:system} explosion image of Uni350.}
%   \vspace{-1.2cm}
% \end{figure}

\section{The UniQuad Series}
\label{sec:system}
The \textbf{UniQuad} series adopts a unified conceptual design along with modular components, providing versatility and flexibility for a wide range of UAV missions. 
Specifically, the conceptual design of the \textbf{UniQuad} series consists of three layers separated according to physical position.
The \textit{upper layer} offers spacious mounting space for various sensors, such as stereo cameras, event cameras, LiDARs and GPS units.
This layer also contains an NVIDIA Jetson Orin NX\footnote{\url{https://www.nvidia.com/en-us/autonomous-machines/embedded-systems/jetson-orin}} for high-level computational tasks, which can be substituted with other computing units such as Intel NUC, Raspberry Pi, LubanCat, etc.
The \textit{middle layer} holds the propulsion system driven by four brushless motors and the low-level flight controller Nxt-FC\footnote{\url{https://github.com/HKUST-Aerial-Robotics/Nxt-FC}}.
The \textit{lower layer} can optionally carry additional payloads, such as gimbals or manipulators.
In this article, we provide four examples of \textbf{UniQuad} series, namely \textbf{Uni127C}, \textbf{Uni250C}, \textbf{Uni250L} and \textbf{Uni350CL}, with specifications listed in Table \ref{tab:exp_scenarios}.
For the detailed bill of material (BOM) and 3D CAD design files, please refer to the project page\footnote{\url{https://hkust-aerial-robotics.github.io/UniQuad/}}.

\section{Experiments}
\label{sec:exp}

\begin{table*}[t]
  \caption{Specification of UniQuad Series}
  % \vspace{-0.2cm}
    % \centering
  % \begin{tabular}{M{0.9cm}M{0.9cm}M{0.9cm}M{0.9cm}M{1.15cm}M{1.55cm}}
  \begin{tabular}{cccccccccc}
    \toprule\toprule
    \textbf{Model} & \textbf{Wheelbase} & \textbf{Motor$^\dagger$} & \textbf{Propeller} & \textbf{Weight} & \begin{tabular}[c]{@{}c@{}}\textbf{Battery}\\\textbf{Capacity}\end{tabular} & \textbf{Equipped Sensor} & \textbf{TWR$^\ddagger$} & \begin{tabular}[c]{@{}c@{}}\textbf{Flight}\\\textbf{Duration}\end{tabular} \\
    \midrule
    \textbf{Uni127C} & 127mm & F1404 KV3800 & Gemfan 3016 & 468g & 4S 3000mAh 30C & Intel RealSense D430 & $\approx$ 2.77  & 7min \\ [0.2cm]
    \textbf{Uni250C} & 250mm & F60PRO KV2550 & Gemfan 5043 & 848g & 4S 2200mAh 45C & Intel RealSense D435 & $\approx$ 7.03 & 8.5min \\ [0.2cm]
    \textbf{Uni250L} & 250mm & F60PRO KV1750 & Gemfan 5043 & 1258g  & 6S 3300mAh 50C & Livox LiDAR Mid-360 & $\approx$ 5.46 & 12min \\ [0.2cm]
    \textbf{Uni350CL} & 350mm & F90 KV1300 & Sunnysky EOLO8 & 1526g & 6S 3300mAh 50C & \begin{tabular}[c]{@{}c@{}} D435 \& Mid-360 \end{tabular} & $\approx$ 5.54 & 15.5min \\ 

    \toprule\toprule
  \end{tabular} \\
  \footnotesize{$^\dagger$ All motors are purchased from T-MOTOR (\url{https://store.tmotor.com/categorys/f-series-motor})}\\
  \footnotesize{$^\ddagger$ TWR: Thrust-to-Weight Ratio. The value is calculated using data obtained from the motor test report available on the T-MOTOR official website.}\\
  \label{tab:exp_scenarios}
  \vspace{-0.6cm}
\end{table*}
 
\begin{figure*}[t]
	\centering
  \subfigtopskip=0pt
	\subfigbottomskip=2pt
	\subfigcapskip=-3pt
  \subfigure[\textbf{Uni127C, e=(0.04,0.02,0.03)}]{\includegraphics[width=0.5\columnwidth]{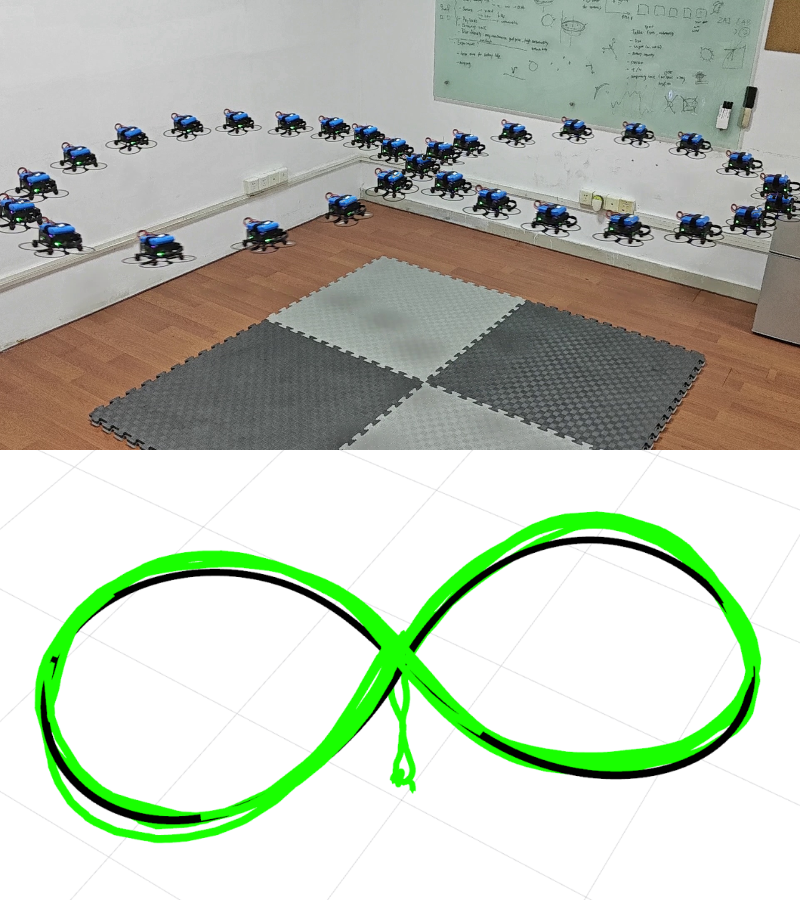}}
  \subfigure[\textbf{Uni250C, e = (0.02,0.04,0.01)}]{\includegraphics[width=0.5\columnwidth]{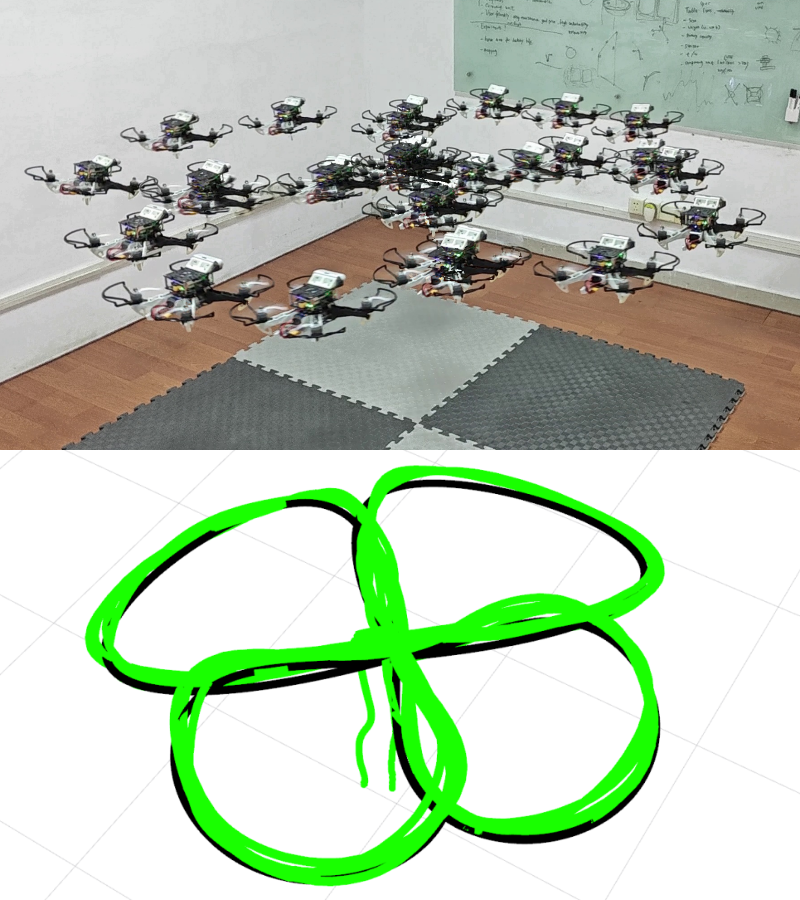}}
  \subfigure[\textbf{Uni250L, e = (0.02,0.04,0.03)}]{\includegraphics[width=0.5\columnwidth]{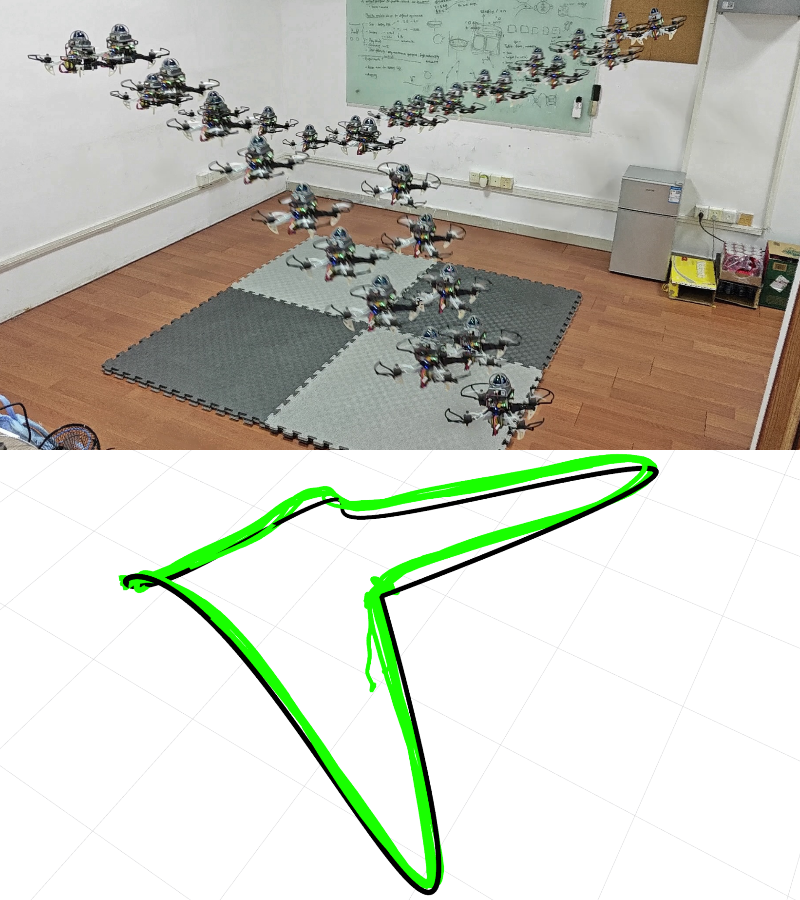}}
  \subfigure[\textbf{Uni350CL, e = (0.04,0.04,0.01)}]{\includegraphics[width=0.5\columnwidth]{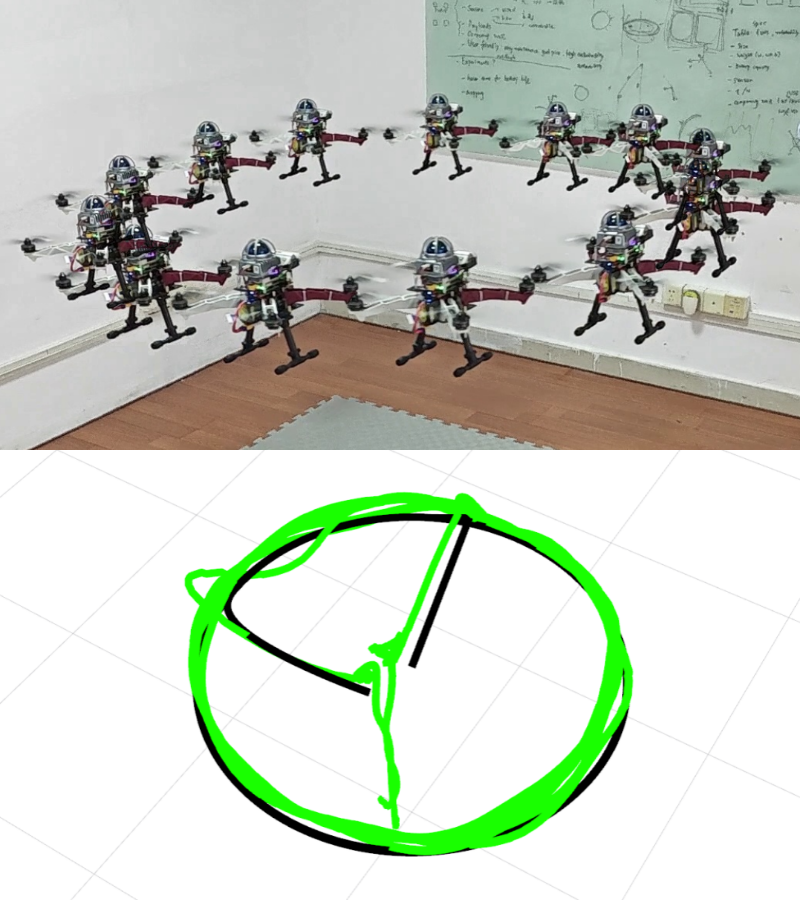}} 
  % \subfigure{\includegraphics[width=0.39\columnwidth]{cat_think.jpg}}
  % \vspace{-0.2cm}
  \caption{\label{fig:exp_traj} Experiment results of the trajectory following experiments. The first row shows the composite images of the flight recordings. The second row illustrates the trajectory commands in black curves and the onboard state estimations in green curves. The average tracking errors(\(m\)) during the experiments in (\(x,y,z\)) axes are indicated in each subcaption respectively.}
  \vspace{-0.4cm}
\end{figure*}

% \begin{figure}[t]
% 	\centering
%   \subfigtopskip=0pt
% 	\subfigbottomskip=2pt
% 	\subfigcapskip=-3pt
%   \subfigure[Stereo Camera]{\includegraphics[width=0.45\columnwidth]{cat_think.jpg}} \hskip -3pt      
%   \subfigure[LiDAR]{\includegraphics[width=0.45\columnwidth]{cat_think.jpg}} \hskip -3pt
% %   \vspace{-0.1cm}
%   \caption{\label{fig:exp_mapping} exp of mapping}
% %   \vspace{-0.8cm}
% \end{figure}

To validate the performance of the \textbf{UniQuad} series, we conduct trajectory following experiments, which is one of the fundamentals for UAV autonomous flights.
During the experiments, all four \textbf{UniQuad} models are commanded to execute predefined trajectories without relying on external infrastructure such as motion capture systems or GPS.
The onboard state estimation is provided by VINS\cite{qin2018vins} for models equipped with stereo cameras and Fast-LIO\cite{xu2021fast} for LiDAR.
As shown in Fig. \ref{fig:exp_traj}, the results demonstrate that the \textbf{UniQuad} series successfully follows the trajectories with high accuracy and stability. 
% Note that \textbf{Uni350} is carrying a payload of 3kg during the experiment, which can be substitute to other components depending on user's demand.
% For the environment mapping experiments, the \textbf{Unix} and \textbf{Unix} models, equipped with stereo cameras and LiDAR sensors respectively, demonstrate their capability to generate high-quality 3D maps as shown in Fig.\ref{fig:exp_traj}(e).

% Conclusion
\section{Conclusions}
\label{sec:conclude}
% 70 - 100 words
% This article introduces a fully open-source and versatile quadrotor platform series \textbf{UniQuad}, which shares a unified conceptual design with modular components to provide flexible and customizable solutions for diverse UAV missions.
This article presents the \textbf{UniQuad} series, a fully open-source and versatile quadrotor platform. This series shares a unified conceptual design with modular components, offering flexible and customizable solutions for a wide range of UAV missions. 
% We demonstrated \textbf{UniQuad} series's capability to track predefined trajectories with high accuracy and stability.
We demonstrate the \textbf{UniQuad} series' capability to track predefined trajectories with exceptional accuracy and stability.
For more comprehensive tasks, the \textbf{UniQuad} series has showcased its potential in various autonomous flight projects, such as perception-aware planning \cite{chen2024apace} and autonomous exploration \cite{zhang2024falcon}.
The \textbf{UniQuad} series is expected to serve as a valuable package, paving the path for real-world verifications and applications.

\addtolength{\textheight}{0.cm}   % This command serves to balance the column lengths

\newlength{\bibitemsep}\setlength{\bibitemsep}{0.0\baselineskip}
\newlength{\bibparskip}\setlength{\bibparskip}{0.1pt}
\let\oldthebibliography\thebibliography
\renewcommand\thebibliography[1]{%
\oldthebibliography{#1}%
\setlength{\parskip}{\bibitemsep}%
\setlength{\itemsep}{\bibparskip}%
}

\bibliography{ref}

\begin{thebibliography}{10}
\providecommand{\url}[1]{#1}
\csname url@samestyle\endcsname
\providecommand{\newblock}{\relax}
\providecommand{\bibinfo}[2]{#2}
\providecommand{\BIBentrySTDinterwordspacing}{\spaceskip=0pt\relax}
\providecommand{\BIBentryALTinterwordstretchfactor}{4}
\providecommand{\BIBentryALTinterwordspacing}{\spaceskip=\fontdimen2\font plus
\BIBentryALTinterwordstretchfactor\fontdimen3\font minus
  \fontdimen4\font\relax}
\providecommand{\BIBforeignlanguage}[2]{{%
\expandafter\ifx\csname l@#1\endcsname\relax
\typeout{** WARNING: IEEEtran.bst: No hyphenation pattern has been}%
\typeout{** loaded for the language `#1'. Using the pattern for}%
\typeout{** the default language instead.}%
\else
\language=\csname l@#1\endcsname
\fi
#2}}
\providecommand{\BIBdecl}{\relax}
\BIBdecl

\bibitem{zhang2024falcon}
Y.~Zhang, X.~Chen, C.~Feng, B.~Zhou, and S.~Shen, ``Falcon: Fast autonomous
  aerial exploration using coverage path guidance,'' 2024, under review.

\bibitem{feng2023fc}
C.~Feng, H.~Li, J.~Jiang, X.~Chen, B.~Zhou, and S.~Shen, ``Fc-planner: A
  skeleton-guided planning framework for fast aerial coverage of complex 3d
  scenes,'' \emph{arXiv preprint arXiv:2309.13882}, 2023.

\bibitem{li2023autotrans}
H.~Li, H.~Wang, C.~Feng, F.~Gao, B.~Zhou, and S.~Shen, ``Autotrans: A complete
  planning and control framework for autonomous uav payload transportation,''
  \emph{IEEE Robotics and Automation Letters}, vol.~8, no.~10, pp. 6859--6866,
  2023.

\bibitem{sa2017build}
I.~Sa, M.~Kamel, M.~Burri, M.~Bloesch, R.~Khanna, M.~Popovi{\'c}, J.~Nieto, and
  R.~Siegwart, ``Build your own visual-inertial drone: A cost-effective and
  open-source autonomous drone,'' \emph{IEEE Robotics \& Automation Magazine},
  vol.~25, no.~1, pp. 89--103, 2017.

\bibitem{mohta2018fast}
K.~Mohta, M.~Watterson, Y.~Mulgaonkar, S.~Liu, C.~Qu, A.~Makineni, K.~Saulnier,
  K.~Sun, A.~Zhu, J.~Delmerico \emph{et~al.}, ``Fast, autonomous flight in
  gps-denied and cluttered environments,'' \emph{Journal of Field Robotics},
  vol.~35, no.~1, pp. 101--120, 2018.

\bibitem{baca2021mrs}
T.~Baca, M.~Petrlik, M.~Vrba, V.~Spurny, R.~Penicka, D.~Hert, and M.~Saska,
  ``The mrs uav system: Pushing the frontiers of reproducible research,
  real-world deployment, and education with autonomous unmanned aerial
  vehicles,'' \emph{Journal of Intelligent \& Robotic Systems}, vol. 102,
  no.~1, p.~26, 2021.

\bibitem{foehn2022agilicious}
P.~Foehn, E.~Kaufmann, A.~Romero, R.~Penicka, S.~Sun, L.~Bauersfeld,
  T.~Laengle, G.~Cioffi, Y.~Song, A.~Loquercio \emph{et~al.}, ``Agilicious:
  Open-source and open-hardware agile quadrotor for vision-based flight,''
  \emph{Science robotics}, vol.~7, no.~67, p. eabl6259, 2022.

\bibitem{zhou2022swarm}
X.~Zhou, X.~Wen, Z.~Wang, Y.~Gao, H.~Li, Q.~Wang, T.~Yang, H.~Lu, Y.~Cao, C.~Xu
  \emph{et~al.}, ``Swarm of micro flying robots in the wild,'' \emph{Science
  Robotics}, vol.~7, no.~66, p. eabm5954, 2022.

\bibitem{pan2023canfly}
N.~Pan, R.~Jin, C.~Xu, and F.~Gao, ``Canfly: A can-sized autonomous mini
  coaxial helicopter,'' in \emph{2023 IEEE/RSJ International Conference on
  Intelligent Robots and Systems (IROS)}.\hskip 1em plus 0.5em minus
  0.4em\relax IEEE, 2023, pp. 4989--4996.

\bibitem{liu2024omninxt}
P.~Liu, C.~Feng, Y.~Xu, Y.~Ning, H.~Xu, and S.~Shen, ``Omninxt: A fully
  open-source and compact aerial robot with omnidirectional visual
  perception,'' \emph{arXiv preprint arXiv:2403.20085}, 2024.

\bibitem{qin2018vins}
T.~Qin, P.~Li, and S.~Shen, ``Vins-mono: A robust and versatile monocular
  visual-inertial state estimator,'' \emph{{IEEE} Trans. Robot. ({TRO})},
  vol.~34, no.~4, pp. 1004--1020, 2018.

\bibitem{xu2021fast}
W.~Xu and F.~Zhang, ``Fast-lio: A fast, robust lidar-inertial odometry package
  by tightly-coupled iterated kalman filter,'' \emph{IEEE Robotics and
  Automation Letters}, vol.~6, no.~2, pp. 3317--3324, 2021.

\bibitem{chen2024apace}
X.~Chen, Y.~Zhang, B.~Zhou, and S.~Shen, ``Apace: Agile and perception-aware
  trajectory generation for quadrotor flights,'' \emph{arXiv preprint
  arXiv:2403.08365}, 2024.

\end{thebibliography}

\end{document}